\documentclass[11pt]{article}

\usepackage[preprint]{acl}

\usepackage{times}
\usepackage{latexsym}
\usepackage{booktabs}
\usepackage{multirow}
\usepackage{array}
\usepackage{amsmath}
\usepackage{amssymb}
\usepackage{float}    
\usepackage{graphicx} 
\usepackage{caption}  
\usepackage{enumitem} 

\usepackage[T1]{fontenc}

\usepackage[utf8]{inputenc}

\usepackage{microtype}

\usepackage{inconsolata}

\usepackage{graphicx}

\title{Expert-Guided Multimodal Fusion for Unified Emotion and Sentiment Analysis}

\author{Jiaqi Qiao \quad Xinran Li \quad Yifan Lyu \quad Xiujuan Xu \quad Liu Yu \\
  Dalian University of Technology, Dalian, China}

\begin{document}
\maketitle
\begin{abstract}
Multimodal emotion understanding requires the integration of heterogeneous data sources, including text, audio, and visual modalities, while simultaneously addressing discrete emotion recognition and continuous sentiment analysis. We propose EGMF, a unified framework that combines expert-guided multimodal fusion with large language models to achieve superior performance across both tasks. At the core of our framework is a multi-scale expert network, comprising a local expert for capturing subtle emotional nuances, a semantic correlation expert for modeling cross-modal relationships, and a global context expert for understanding long-range dependencies. These experts are adaptively integrated via hierarchical dynamic gating, enabling context-aware feature selection and modality weighting. The enhanced multimodal representations are seamlessly incorporated into the language model through pseudo token injection and prompt-based conditioning, allowing a single generative framework to handle both classification and regression tasks. We employ parameter-efficient LoRA fine-tuning to maintain computational efficiency. Extensive experiments on bilingual benchmark datasets (MELD, CHERMA, MOSEI, SIMS-V2) demonstrate that EGMF outperforms state-of-the-art methods in terms of accuracy, cross-lingual robustness, and the discovery of universal patterns in multimodal emotional expressions.
\end{abstract}

\section{Introduction}

Understanding human emotions from multimodal signals, including text, audio, and visual cues, is a central goal in affective computing. Emotion Recognition in Conversation (ERC)~\citep{1.gkintoni2025neural} and Multimodal Sentiment Analysis (MSA)~\cite{2.yang2025ccin} play critical roles in applications such as mental health assessment~\cite{3.benrouba2023emotional}, human-computer interaction, and social media analysis~\cite{4.rodriguez2023review}. However, the heterogeneity of modalities, the complexity of cross-modal interactions, and the semantic gap between low-level perception and high-level emotional reasoning pose significant challenges for robust and generalizable emotion understanding.

Large Language Models (LLMs) exhibit strong multi-task generalization and contextual reasoning capabilities, offering new opportunities for multimodal emotion understanding~\cite{5.chen2024evolution}. Existing approaches often use LLMs merely as classifiers or integrate multimodal inputs via simple concatenation~\cite{6.Nguyen_2023}, underutilizing their cross-modal reasoning potential. Traditional methods for conversational emotion understanding also face limitations: RNN-based methods, such as DS-LSTM~\citep{11.DS-LSTM} and DialogueCRN~\citep{12.ghosal2019dialoguegcn}, handle temporal context but struggle with long-range dependencies; Transformer-based methods, including EmoBERTa~\citep{15.kim2021emoberta} and BERT-ERC~\citep{18.qin2023bert}, perform well for single-modal representation but rely on static fusion mechanisms that cannot dynamically adjust modality importance; graph neural network approaches, such as DialogueGCN~\citep{24.hu2021dialoguecrn} and DAG-ERC~\citep{26.shen2021directed}, model dialogue relations via graph structures but are constrained by fixed topologies and limited edge designs.

To address these challenges, we propose an \textbf{adaptive expert-guided multimodal fusion framework}, introducing three functionally specialized expert networks to handle fine-grained local features, semantic correlation patterns, and global contextual information. These experts are integrated via \textbf{hierarchical dynamic gating} for context-aware feature selection. Enhanced multimodal representations are combined with the generative capabilities of LLMs to effectively model long-range dependencies and complex dialogue contexts. We further employ \textbf{parameter-efficient LoRA fine-tuning} to reduce computational cost while maintaining performance.

Extensive experiments on Chinese and English datasets demonstrate that our method outperforms existing baselines in accuracy, cross-lingual adaptability, and computational efficiency. Our main contributions are threefold:
\begin{itemize}
    \item We propose a unified multimodal emotion understanding framework, supporting both ERC classification and MSA regression within a single architecture;
    \item We design an adaptive feature enhancement module based on multi-scale expert networks and hierarchical dynamic gating, significantly improving multimodal representation expressiveness;
    \item We achieve state-of-the-art performance on multiple Chinese and English datasets, demonstrating strong cross-lingual generalization and establishing a new paradigm for unified multimodal affective computing.
\end{itemize}

\section{Related Work}
ERC and MSA are two core tasks in affective computing that rely on the effective fusion of heterogeneous modalities, including text, audio, and visual signals. While MSA typically focuses on polarity classification at the utterance level, ERC emphasizes the modeling of emotional dynamics and contextual dependencies across multi-turn dialogues. Recent research has increasingly explored large language models (LLMs) for these tasks, in addition to traditional approaches based on RNNs, Transformers, and GNNs.

\paragraph{LLM-based Methods.}
With their powerful pretrained knowledge and contextual reasoning abilities, LLMs have been introduced into emotion recognition tasks. InstructERC \citep{8.lei2023instructerc} first reformulates ER as a generative task, guiding LLMs to produce emotion labels via prompt-based learning, thereby improving generalization across domains. However, this approach relies solely on textual inputs and does not incorporate multimodal information or distinguish between task types such as classification and regression.To address these limitations, BiosERC \citep{9.xue2024bioserc} incorporates speaker biography information into ERC and leverages LLMs to extract background knowledge of speakers, enhancing contextual emotional understanding. DialogueLLM \citep{10.zhang2023dialoguellm} further integrates visual and textual inputs and applies instruction tuning for multimodal sentiment classification, demonstrating promising adaptability in complex dialog scenarios.

Although these studies reveal the potential of LLMs in affective computing, existing methods still face several critical limitations. Most approaches target either SA or ER exclusively, lacking unified modeling capabilities. Moreover, multimodal input designs are typically static and fail to dynamically adapt to speaker shifts and modality dependencies, while the lack of structured fusion interfaces limits effective utilization of non-text modalities. To address these limitations, we propose a unified framework that integrates expert-guided fusion strategies with a generative LLM backbone, enabling flexible and generalizable modeling for both ER and SA tasks.

For more detailed discussions of RNN, Transformer, and GNN-based methods, please refer to Appendix \ref{appendix:baselines}.

\section{Method}
\subsection{Task Definition}
Given a multimodal dialogue sequence containing $N$ utterances, each utterance $u_i$ involves aligned data sources from three modalities: textual (t), acoustic (a), and visual (v). This relationship can be represented as $\mathcal{X} = \{\mathcal{X}^t, \mathcal{X}^a, \mathcal{X}^v\}$, where $\mathcal{X}^m = \{u^m_1, u^m_2, ..., u^m_N\}$ and $m \in \{t, a, v\}$. Each utterance representation for modality $m$ is denoted as $u^m_i \in \mathbb{R}^{L_m \times d_m}$, where $L_m$ and $d_m$ represent the sequence length and feature dimension of modality $m$, respectively.

Our goal is to design a mapping function $F$ that transforms the multimodal input $\mathcal{X}$ into emotional prediction results. The framework supports two task paradigms:

\paragraph{Emotion Recognition.} We map multimodal input to a discrete set of emotion categories $\mathcal{E} = \{e_1, e_2, ..., e_K\}$, where $K$ denotes the number of emotion classes. The mapping relationship is defined as $F_{cls}: \mathcal{X} \rightarrow \mathcal{E}$. Specifically, given input $\mathcal{X}$, the model predicts the emotion category $\hat{y} = \arg\max_{e_k \in \mathcal{E}} P(e_k|\mathcal{X})$.

\paragraph{Sentiment Analysis.} We map multimodal input to a continuous sentiment intensity space $\mathcal{S} \subset \mathbb{R}$, typically defined within a bounded interval $[s_{min}, s_{max}]$. The mapping relationship is defined as $F_{reg}: \mathcal{X} \rightarrow \mathcal{S}$. Specifically, given input $\mathcal{X}$, the model predicts the sentiment intensity $\hat{s} = \mathbb{E}[s|\mathcal{X}]$, where $s \in \mathcal{S}$ represents the true sentiment intensity value.

\begin{figure*}[t]
\centering
\includegraphics[width=0.9\textwidth]{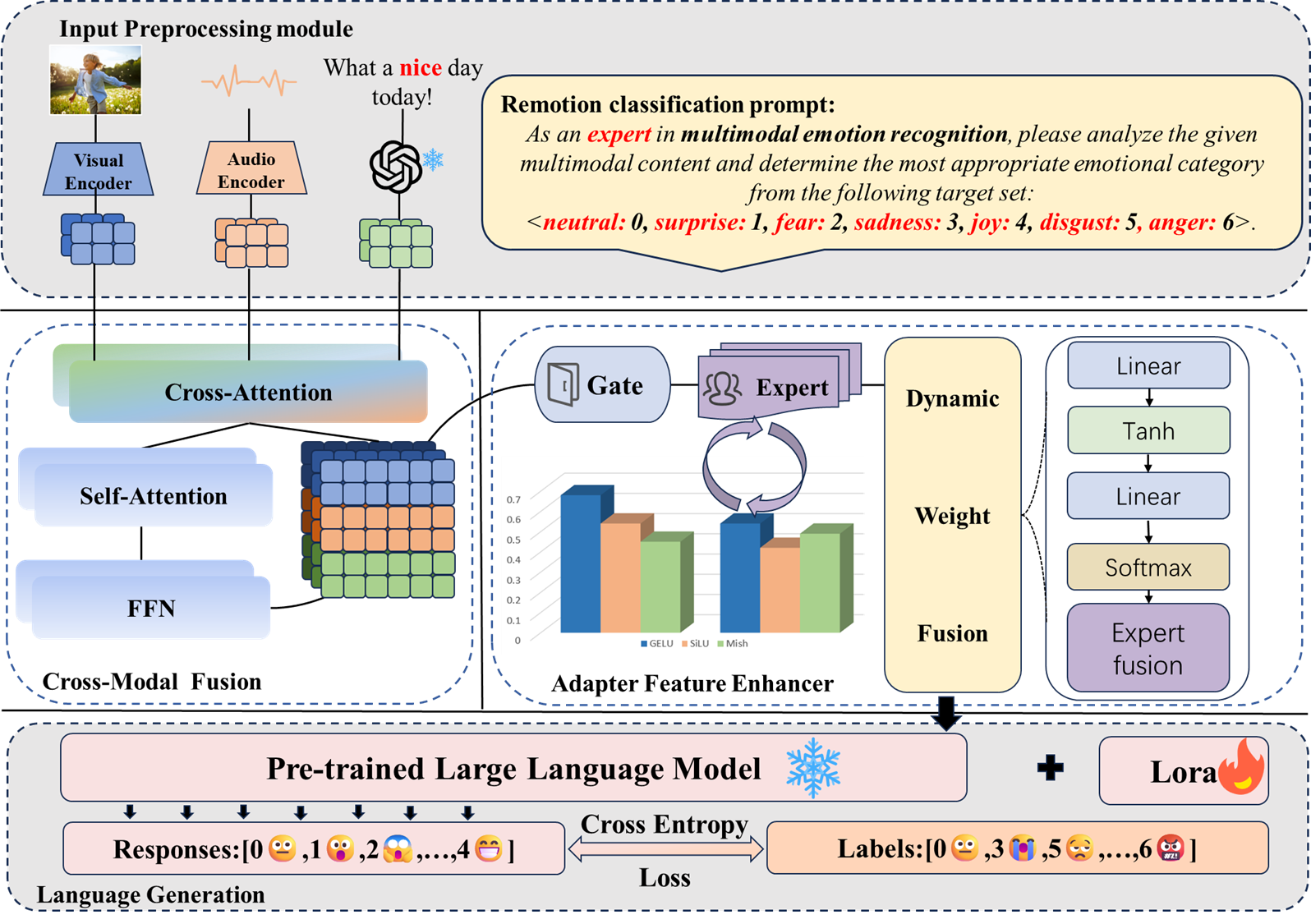}
\caption{Architecture of the proposed EGMF (Expert-Guided Multimodal Fusion) framework for unified emotion recognition and sentiment analysis. The model integrates expert networks with large language models, featuring Transformer-based multimodal encoders, cross-modal attention fusion, adaptive feature enhancement, and LoRA-tuned generative prediction.}
\label{fig:egmf_overview}
\end{figure*}

\subsection{Overall Framework}
Our proposed EGMF (Expert-Guided Multimodal Fusion) framework addresses the challenge of unified emotion recognition and sentiment analysis through a novel integration of expert-guided fusion mechanisms with large language models. As illustrated in Figure \ref{fig:egmf_overview}, the framework consists of four core modules that work synergistically to transform multimodal inputs into emotional predictions.

The \textbf{Input Preprocessing Module} transforms raw multimodal data into unified feature representations, while the \textbf{Cross-Modal Fusion Module} captures inter-modal dependencies through cross-modal Transformer architecture.

The \textbf{Adaptive Feature Enhancer Module} represents the key innovation of our approach. This module employs three functionally specialized expert networks: a fine-grained local expert for capturing subtle emotional nuances, a semantic correlation expert for modeling cross-modal relationships, and a global context expert for understanding long-range dependencies. These experts operate with different bottleneck architectures and activation functions to capture complementary aspects of emotional semantics. The module incorporates a hierarchical dynamic gating mechanism that performs two-stage adaptive weighting: initial feature-driven gating based on input characteristics, followed by context-aware adjustment. This design enables the framework to dynamically adjust the contribution of each expert based on contextual requirements, generating enhanced multimodal representations that are significantly more discriminative for emotion understanding tasks.

The \textbf{Language Generation Module} integrates the enhanced features with a large language model through LoRA fine-tuning, enabling unified generation-based prediction for both classification and regression tasks.

\subsection{Input Preprocessing Module}
The Input Preprocessing Module transforms heterogeneous multimodal data into unified feature representations, serving as the foundation for subsequent cross-modal fusion. Given the raw multimodal input $\mathcal{X} = \{\mathcal{X}^t, \mathcal{X}^a, \mathcal{X}^v\}$, this module processes each modality through specialized encoders to generate compatible feature spaces.

\subsubsection{Audio-Visual Encoders}
For audio and visual modalities, we employ dedicated Transformer-based encoders to capture temporal dependencies. Given an input sequence $u^m_i \in \mathbb{R}^{L_m \times d_m}$ for modality $m \in \{a, v\}$, the encoding process involves feature projection, positional encoding, multi-layer Transformer encoding, and masked global average pooling to handle variable-length sequences: $\mathbf{f}^m_i = \text{AudioVisualEncoder}(u^m_i) \in \mathbb{R}^{d_{av}}$, where $d_{av} = 256$ for both audio and visual modalities.

\subsubsection{Text Embedding via Large Language Models}
For textual modality, we leverage the embedding layer of pre-trained large language models to capture rich semantic representations: $\mathbf{f}^t_i = \text{LLM}_{\text{embed}}(\mathcal{X}^t_i) \in \mathbb{R}^{L_t \times d_{emb}}$, where $d_{emb}$ represents the embedding dimension of the specific language model.

\subsubsection{Unified Feature Representation}
The preprocessing module outputs unified representations for all modalities: $\mathcal{F} = \{\mathbf{f}^t_1, ..., \mathbf{f}^t_N, \mathbf{f}^a_1, ..., \mathbf{f}^a_N, \mathbf{f}^v_1, ..., \mathbf{f}^v_N\}$. These representations ensure compatibility for subsequent cross-modal fusion while preserving modality-specific characteristics.

\subsection{Cross-Modal Fusion Module}
\subsubsection{Feature Projection and Alignment}
Given the preprocessed features $\mathbf{f}^t_i \in \mathbb{R}^{L_t \times d_{emb}}$, $\mathbf{f}^a_i \in \mathbb{R}^{d_{av}}$, and $\mathbf{f}^v_i \in \mathbb{R}^{d_{av}}$, we first project all modalities to a unified hidden dimension $d_h$ to ensure compatibility for cross-modal attention. Each modality undergoes linear transformation $\mathbf{H}^m_i = \text{Linear}_m(\mathbf{f}^m_i)$ for $m \in \{t, a, v\}$, resulting in aligned representations $\mathbf{H}^m_i \in \mathbb{R}^{L_m \times d_h}$, where $d_h = 512$ represents the unified hidden dimension for cross-modal interaction.

\subsubsection{Bidirectional Cross-Modal Attention}
We construct combined audio-visual representations by concatenating the aligned audio and visual features, yielding $\mathbf{H}^{av}_i \in \mathbb{R}^{2 \times d_h}$, and apply bidirectional cross-attention mechanisms to capture inter-modal dependencies. The cross-modal attention mechanism uses text as query and audio-visual features as key and value:
\begin{equation}
\small
\mathbf{Z}^{cross}_i = \text{CrossAttention}(\mathbf{H}^t_i, \mathbf{H}^{av}_i, \mathbf{H}^{av}_i) + \mathbf{H}^t_i
\label{eq:cross_attention}
\end{equation}
\begin{equation}
\small
\mathbf{Z}^{self}_i = \text{SelfAttention}(\mathbf{Z}^{cross}_i, \mathbf{Z}^{cross}_i, \mathbf{Z}^{cross}_i) + \mathbf{Z}^{cross}_i
\label{eq:self_attention}
\end{equation}

\subsubsection{Multimodal Feature Fusion}
The attended features are processed through a feed-forward network and then aggregated through global pooling to obtain the final fused representation:
\begin{equation}
\mathbf{Z}^{ffn}_i = \text{FFN}(\mathbf{Z}^{self}_i) + \mathbf{Z}^{self}_i
\label{eq:ffn_processing}
\end{equation}
\begin{equation}
\mathbf{f}^{fusion}_i = \text{GlobalPool}(\mathbf{Z}^{ffn}_i) \in \mathbb{R}^{d_h}
\label{eq:global_pooling}
\end{equation}
where the global pooling operation aggregates the sequential information into a fixed-dimensional representation for subsequent processing. The cross-modal fusion module thus transforms the heterogeneous multimodal inputs into a unified representation $\mathbf{f}^{fusion}_i$ that captures both intra-modal characteristics and inter-modal interactions, serving as the foundation for adaptive feature enhancement.

\subsection{Adaptive Feature Enhancer Module}

\subsubsection{Multi-Scale Expert Networks Architecture}
We design three functionally specialized expert networks to capture complementary aspects of multimodal emotional semantics. The varying bottleneck dimensions implement different levels of information compression, enabling extraction from fine-grained details to global contextual patterns:

\begin{enumerate}
    \item \textbf{Fine-Grained Local Expert ($E_1$)}: Uses compact bottleneck (ratio 1:8) with Mish activation to capture local details and subtle emotional nuances:
    \begin{equation}
    \mathbf{e}_1^i = E_1(\mathbf{f}^{fusion}_i; \theta_{d_h/8}, \text{Mish}) \in \mathbb{R}^{d_h}
    \label{eq:expert_1}
    \end{equation}
    
    \item \textbf{Semantic Correlation Expert ($E_2$)}: Employs moderate bottleneck (ratio 1:4) with GELU activation for mid-level semantic patterns and cross-modal correlations:
    \begin{equation}
    \mathbf{e}_2^i = E_2(\mathbf{f}^{fusion}_i; \theta_{d_h/4}, \text{GELU}) \in \mathbb{R}^{d_h}
    \label{eq:expert_2}
    \end{equation}
    
    \item \textbf{Global Context Expert ($E_3$)}: Utilizes expanded bottleneck (ratio 1:2) with Swish activation for global context and long-range dependencies:
    \begin{equation}
    \mathbf{e}_3^i = E_3(\mathbf{f}^{fusion}_i; \theta_{d_h/2}, \text{Swish}) \in \mathbb{R}^{d_h}
    \label{eq:expert_3}
    \end{equation}
\end{enumerate}

\subsubsection{Hierarchical Dynamic Gating Mechanism}
We introduce a two-stage hierarchical gating strategy for dynamic expert contribution modulation:

\begin{enumerate}
    \item \textbf{Feature-Driven Gating}: Initial expert preferences based on input characteristics:
    \begin{equation}
    \mathbf{w}_i = \text{GateNetwork}(\mathbf{f}^{fusion}_i) \in \mathbb{R}^3
    \label{eq:feature_gating}
    \end{equation}
    
    \item \textbf{Context-Aware Adjustment}: Refined weighting considering both input features and initial gating:
    \begin{equation}
    \mathbf{c}_i = \text{Concat}(\mathbf{f}^{fusion}_i, \mathbf{w}_i) \in \mathbb{R}^{d_h + 3}
    \label{eq:context_concat}
    \end{equation}
    \begin{equation}
    \boldsymbol{\alpha}_i = \text{Softmax}(\text{MLP}^{dynamic}(\mathbf{c}_i)) \in \mathbb{R}^3
    \label{eq:dynamic_weighting}
    \end{equation}
\end{enumerate}

\subsubsection{Adaptive Weighted Fusion}
The enhanced representation integrates expert outputs through adaptive weighting with learned residual connection:
\begin{equation}
\mathbf{f}^{enhanced}_i = \sum_{k=1}^{3} \alpha_{i,k} \cdot \mathbf{e}_k^i + \beta_i \cdot \mathbf{f}^{fusion}_i
\label{eq:adaptive_fusion}
\end{equation}
where the first term represents the dynamically weighted combination of the three expert outputs ($\mathbf{e}_1^i$, $\mathbf{e}_2^i$, $\mathbf{e}_3^i$) using context-aware weights $\alpha_{i,k}$, and the second term provides a residual connection that preserves the original fused features $\mathbf{f}^{fusion}_i$ through a learned weighting parameter $\beta_i$.

\subsubsection{Pseudo Token Generation}
Enhanced features are transformed into pseudo tokens for LLM integration. These pseudo tokens serve as ``multimodal translators'' that encode rich emotional semantics in a format interpretable by language models, enabling seamless multimodal-to-linguistic feature translation:
\begin{equation}
\mathbf{P}_i = \text{Linear}_{proj}(\mathbf{f}^{enhanced}_i) \in \mathbb{R}^{d_{emb}}
\label{eq:token_projection}
\end{equation}
\begin{equation}
\mathbf{T}^{pseudo}_i = \text{Repeat}(\mathbf{P}_i, n_{tokens}) \in \mathbb{R}^{n_{tokens} \times d_{emb}}
\label{eq:token_repeat}
\end{equation}

\subsection{Language Generation Module}
\subsubsection{Multimodal Feature Integration}

The enhanced pseudo tokens $\mathbf{T}^{pseudo}_i$ are seamlessly integrated with text embeddings to form the complete input sequence for the language model:
\begin{equation}
\mathbf{I}^{complete}_i = [\mathbf{f}^t_i; \mathbf{T}^{pseudo}_i] \in \mathbb{R}^{(L_t + n_{tokens}) \times d_{emb}}
\label{eq:complete_input}
\end{equation}
To facilitate multimodal understanding, we employ a specialized prompt wrapping strategy that embeds the multimodal information within structured prompt templates:
\begin{equation}
\mathbf{I}^{wrapped}_i = [\mathbf{E}_{prefix}; \mathbf{T}^{pseudo}_i; \mathbf{E}_{suffix}; \mathbf{P}_{task}]
\label{eq:wrapped_input}
\end{equation}
where $\mathbf{E}_{prefix}$ and $\mathbf{E}_{suffix}$ are embeddings of language-specific prompt templates, and $\mathbf{P}_{task}$ represents task-specific prompts for classification or regression.

\subsubsection{LoRA-based Parameter-Efficient Fine-tuning}
To adapt the pre-trained language model for multimodal emotion understanding while maintaining computational efficiency, we employ Low-Rank Adaptation (LoRA) with the following configuration:
\begin{equation}
\mathbf{W}_{adapted} = \mathbf{W}_{frozen} + \mathbf{B} \mathbf{A}
\label{eq:lora_adaptation}
\end{equation}
where $\mathbf{B} \in \mathbb{R}^{d \times r}$, $\mathbf{A} \in \mathbb{R}^{r \times d}$ with rank $r = 8$, LoRA alpha $\alpha = 16$, and dropout rate $p = 0.1$. The adaptation targets the query-key-value projection matrices of the attention mechanism, enabling effective multimodal adaptation with minimal parameter overhead.

\subsubsection{Unified Generation Framework}
The language model processes the wrapped multimodal input to generate task-specific outputs through conditional text generation:
\begin{equation}
P(y|\mathbf{I}^{wrapped}_i) = \text{LLM}(\mathbf{I}^{wrapped}_i; \theta_{frozen}, \theta_{LoRA})
\label{eq:conditional_generation}
\end{equation}
For training, the model is optimized using standard causal language modeling loss:
\begin{equation}
\mathcal{L} = -\sum_{t} \log P(y_t | \mathbf{I}^{wrapped}_i, y_{<t})
\label{eq:causal_loss}
\end{equation}
where the loss is computed only on the target tokens corresponding to emotion labels or sentiment values.

\subsubsection{Task-Adaptive Output Processing}

During inference, the generated text outputs are processed according to specific task requirements. For emotion recognition (ERC), the model generates discrete emotion category labels that are directly mapped to class indices, while for sentiment analysis (MSA), the model generates numerical values that are parsed into floating-point scores with robust handling of various text formats. This unified generation approach enables our framework to seamlessly handle both classification and regression tasks within a single architecture, leveraging the reasoning capabilities of large language models for improved multimodal emotion understanding.

\section{Datasets}

\begin{table}[h]
\centering
\small
\caption{Dataset Statistics and Task Information}
\label{tab:dataset_stats}
\begin{tabular}{lccccc}
\toprule
\textbf{Dataset} & \textbf{Language}  &  \textbf{Task} & \textbf{Train} & \textbf{Valid} & \textbf{Test}  \\
\midrule
MELD & English & ERC & 9,989 & 1,109 & 2,610   \\
MOSEI & English & MSA & 16,326 & 1,871 & 4,659   \\
CHERMA & Chinese & ERC & 17,230 & 5,743 & 5,744  \\
SIMS-V2 & Chinese & MSA  & 2,722 & 647 & 1,034   \\

\bottomrule
\end{tabular}
\end{table}
We evaluate our EGMF framework on four widely-used multimodal emotion datasets, covering both classification and regression tasks in English and Chinese languages.

\paragraph{MELD}\citep{28.poria-etal-2019-meld} An English emotion recognition dataset extracted from TV series dialogues, featuring multimodal data with seven emotion categories: anger, disgust, fear, joy, neutral, sadness, and surprise. 

\paragraph{CHERMA}\citep{29.sun-etal-2023-layer} A Chinese conversational emotion recognition dataset with seven emotion categories: anger, disgust, fear, happiness, neutral, sadness, and surprise. 

\paragraph{SIMSV2}\citep{30.liu2022make} A Chinese multimodal sentiment analysis dataset designed for regression tasks with sentiment intensity annotations in the range [-1, +1]. 

\paragraph{MOSEI}\citep{31.bagher-zadeh-etal-2018-multimodal} An English multimodal sentiment analysis dataset with sentiment intensity annotations in the range [-3, +3], collected from YouTube videos. 

Table \ref{tab:dataset_stats} provides a detailed comparative analysis of the statistical properties and characteristics across these datasets.

\section{Experiments}

\begin{table*}[h]
\centering
\small
\caption{Performance comparison on MOSEI and SIMS-V2 datasets.}
\begin{tabular}{lcccccccccc}
\toprule
\multirow{2}{*}{Model} & \multicolumn{5}{c}{MOSEI} & \multicolumn{5}{c}{SIMS-V2} \\
\cmidrule(r){2-6} \cmidrule(r){7-11}
& Acc-2 & F1 & Acc-7 & MAE & Corr & Acc-2 & F1 & Acc2 (weak) & MAE & Corr \\
\midrule
UniSA\textsubscript{GPT2} (\citeauthor{58.li2023unisa}) & 71.02 & - & 41.36 & 0.838 & - & - & - & - & - & - \\
MulT (\citeauthor{53.tsai2019multimodal}) & 81.15 & 81.56 & 52.84 & 0.559 & 0.733 & 79.50 & 79.59 & 69.61 & 0.317 & 0.703 \\
MAG-BERT (\citeauthor{52.rahman2020integrating})& 82.51 & 82.77 & 50.41 & 0.583 & 0.741 & 79.79 & 79.78 & 71.87 & 0.334 & 0.691 \\
Self-MM  (\citeauthor{55.yu2021learning})& 82.81 & 82.53 & 53.46 & 0.530 & 0.765 & 79.01 & 78.89 & 71.87 & 0.335 & 0.640 \\
CHFN (\citeauthor{56.zhang2022wenetspeech})& 83.70 & 83.90 & 54.30 & 0.525 & 0.778 & - & - & - & - & - \\
UniSA\textsubscript{T5} (\citeauthor{58.li2023unisa})& 84.22 & - & 52.50 & 0.546 & - & - & - & - & - & - \\
UniSA\textsubscript{BART} (\citeauthor{58.li2023unisa})& 84.93 & - & 50.03 & 0.587 & - & - & - & - & - & - \\
UniMSE (\citeauthor{57.hu2022unimse})& 85.86 & 85.79 & 54.39 & 0.523 & 0.773 & - & - & - & - & - \\
\midrule
EGMF(GLM3-6B) & \textbf{87.30} & \textbf{87.09} & \textbf{55.38} & \textbf{0.496} & \textbf{0.801} & 81.56 & 81.13 & 73.09 & 0.284 & \textbf{0.733} \\
EGMF(llama2-7B) & 87.16 & 86.97 & 54.73 & 0.500 & 0.796 & 77.04 & 76.93 & 70.85 & 0.364 & 0.579 \\
EGMF(llama3-8B) & 86.75 & 86.58 & 47.83 & 0.670 & 0.713 & 57.74 & 42.27 & 63.35 & 0.398 & 0.640 \\
EGMF(GLM4-9B) & 87.08 & 87.00 & 54.78 & 0.514 & 0.790 & \textbf{82.57} & \textbf{82.43} & \textbf{74.70} & \textbf{0.284} & 0.720 \\
\bottomrule
\end{tabular}
\label{tab:mosei-sims-results}
\end{table*}

\subsection{Experimental Setup}

All experiments are repeated five times with different random seeds, and we report the average performance across all runs to ensure statistical reliability. For ERC tasks, we report accuracy and weighted F1-score; for MSA tasks, we report binary and multi-class accuracy, MAE, and Pearson correlation. All experiments are conducted on a single NVIDIA A800 GPU. Further implementation details can be found in Appendix \ref{appendix:B}.

\subsection{Main Results}

Tables \ref{tab:mosei-sims-results} and  \ref{tab:meld-cherma-results} present the performance comparison of our EGMF framework against state-of-the-art baselines across four benchmark datasets.

\textbf{Overall Performance.}
Our EGMF framework achieves significant improvements across all tasks. On emotion recognition (ERC), we obtain 65.57\% weighted F1 score on MELD, surpassing the previous best method UniMSE by 0.06\%. On CHERMA, we achieve 73.90\% weighted F1, representing substantial improvements of 3.36\% over LFMIM. For sentiment analysis (MSA), we achieve 87.09\% F1 score on MOSEI, representing improvements of 1.30\% over UniMSE. On SIMS-V2, our best configuration attains 82.43\% F1 score. These improvements are statistically significant across all evaluation metrics, demonstrating the robustness and effectiveness of our approach in both classification and regression tasks.

\textbf{Cross-lingual Performance Analysis.}
The results reveal interesting patterns in cross-lingual multimodal understanding. Our framework demonstrates stronger relative improvements on Chinese datasets (CHERMA: +3.36\% WF1, SIMS-V2: +2.24\% F1) compared to English datasets (MELD: +0.06\% WF1, MOSEI: +1.30\% F1). This suggests that our expert-guided fusion mechanism is particularly effective for languages with different linguistic structures and cultural contexts. The superior performance on Chinese datasets may be attributed to the enhanced multimodal fusion capabilities, which help compensate for potential limitations in cross-lingual semantic understanding.

\begin{figure}[h]
\centering
\includegraphics[width=1.0\columnwidth]{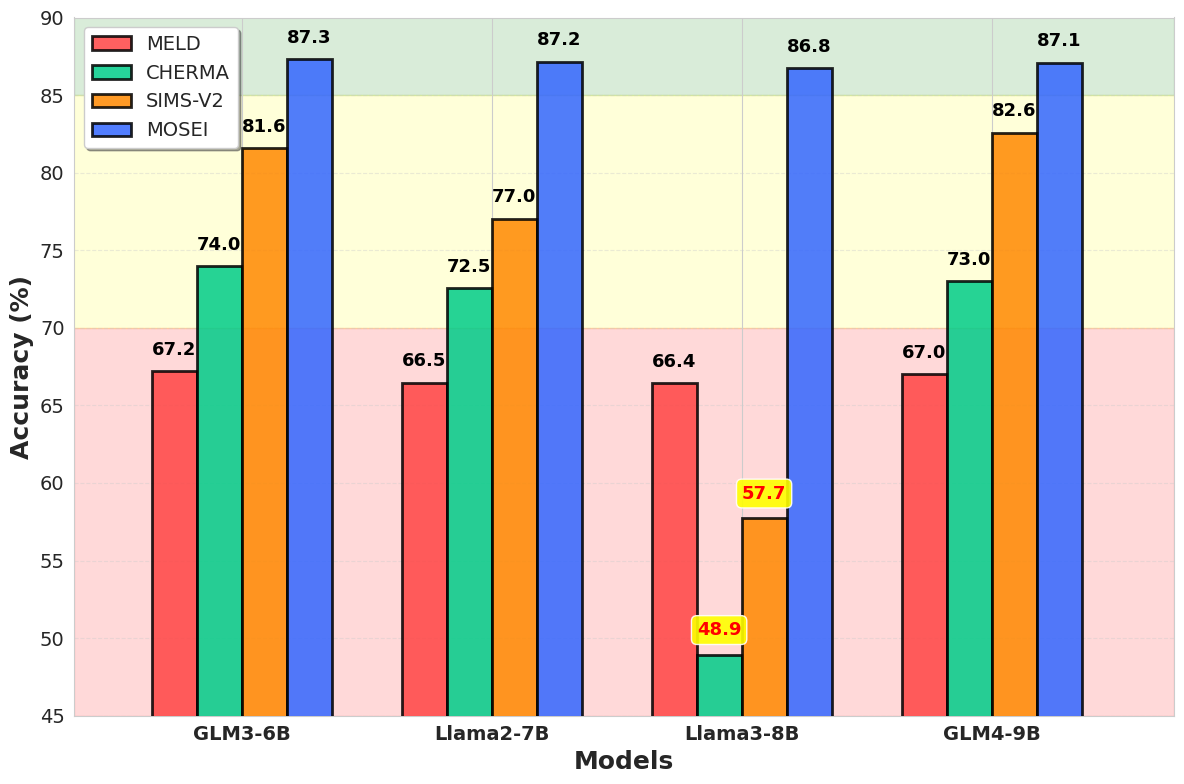}
\caption{Cross-lingual robustness comparison of backbone models on multimodal emotion recognition tasks.}
\label{fig:bar}
\end{figure}

\textbf{Model Configuration Analysis.}
We evaluated four backbone models and selected GLM3-6B\footnote{\url{https://huggingface.co/THUDM/chatglm3-6b}} as our primary configuration. As shown in Figure \ref{fig:bar}, GLM3-6B achieves the best overall performance across both English and Chinese datasets while requiring 33\% fewer parameters than GLM4-9B\footnote{\url{https://huggingface.co/THUDM/glm-4-9b}}. The evaluation also includes Llama2-7B\footnote{\url{https://huggingface.co/meta-llama/Llama-2-7b-hf}}, which demonstrates stable cross-lingual performance, and Llama3-8B\footnote{\url{https://huggingface.co/meta-llama/Meta-Llama-3-8B}}, which exhibits significant performance degradation on Chinese datasets (CHERMA: 46.52\% vs 73.90\% WF1). This cross-lingual stability makes GLM3-6B particularly suitable for multilingual multimodal applications.

The experimental results demonstrate that our framework effectively captures universal emotional patterns while providing an optimal balance between performance and computational efficiency. Additional experimental analyses, including per-class performance breakdown and detailed hyperparameter settings, are provided in Appendix ~\ref{app:detailed_analysis}.

\begin{table}[h]
\centering
\small
\caption{Performance comparison on MELD and CHERMA datasets.}
\begin{tabular}{>{\raggedright\arraybackslash}m{2.5cm}cc!{\vrule}cc}
\toprule
\multirow{2}{*}{Model} & \multicolumn{2}{c}{MELD} & \multicolumn{2}{c}{CHERMA} \\
\cmidrule(r){2-3} \cmidrule(r){4-5}
& Acc & WF1 & Acc & WF1 \\
\midrule
TFN (\citeauthor{50.zadeh2017tensor})& 60.77 & 57.74 & - & 68.37 \\
LMF (\citeauthor{51.liu2018efficient})& 61.15 & 58.30 & - & 68.23 \\
MulT (\citeauthor{53.tsai2019multimodal})& - & - & - & 69.24 \\
PMR (\citeauthor{59.lv2021progressive})& - & - & - & 69.53 \\
LFMIM (\citeauthor{29.sun-etal-2023-layer})& - & - & - & 70.54 \\
GA2MIF (\citeauthor{60.li2023ga2mif})& 61.65 & 58.94 & - & - \\
UniSA\textsubscript{T5} (\citeauthor{58.li2023unisa})& 64.52 & 62.17 & - & - \\
EmoCaps (\citeauthor{17.li2022emocaps})& - & 64.00 & - & - \\
UniMSE(\citeauthor{57.hu2022unimse}) & 65.09 & 65.51 & - & - \\
\midrule
EGMF(GLM3-6B) & \textbf{67.22} & \textbf{65.57} & \textbf{73.97} & \textbf{73.90} \\
EGMF(llama2-7B) & 66.46 & 65.42 & 72.54 & 72.45 \\
EGMF(llama3-8B) & 66.42 & 65.04 & 48.94 & 46.52 \\
EGMF(GLM4-9B) & 67.01 & 65.21 & 73.00 & 73.03 \\
\bottomrule
\end{tabular}
\label{tab:meld-cherma-results}
\end{table}

\begin{table*}[h]
\centering
\small
\caption{Ablation study results on MELD, CHERMA, MOSEI, and SIMS-V2 datasets.}
\setlength{\tabcolsep}{5.5pt}
\begin{tabular}{lcccccccc}
\toprule
\multirow{2}{*}{Model} & \multicolumn{2}{c}{MELD} & \multicolumn{2}{c}{CHERMA} & \multicolumn{2}{c}{MOSEI} & \multicolumn{2}{c}{SIMS-V2} \\
\cmidrule(r){2-3} \cmidrule(r){4-5} \cmidrule(r){6-7} \cmidrule(r){8-9}
& Acc & WF1 & Acc & WF1 & Acc & F1 & Acc & F1 \\
\midrule
w/o A       & 66.70 & 64.64 & 72.98 & 72.77 & 87.29 & 87.10 & 79.98 & 79.24 \\
w/o V       & 66.37 & 64.17 & 72.42 & 72.10 & 87.06 & 86.78 & 78.72 & 77.73 \\
w/o T       & 46.78 & 35.20 & 70.60 & 70.50 & 71.22 & 59.98 & 71.95 & 71.75 \\
w/o A, V    & 61.32 & 61.55 & 57.02 & 56.47 & 86.86 & 86.32 & 79.69 & 79.58 \\
w/o LoRA    & 65.82 & 64.39 & -     & -     & 86.56 & 86.31 & -     & -     \\
w/o Expert($E_1$)     & 66.44 & 65.79 & 73.21 & 72.63 & 85.64 & 85.34 & 80.66 & 80.30 \\
w/o Expert($E_2$)     & 66.13 & 65.55 & 73.03 & 72.85 & 86.91 & 86.71 & 81.24 & 80.95 \\
w/o Expert($E_3$)     & 65.52 & 63.95 & 72.99 & 72.37 & 86.95 & 86.88 & 80.93 & 80.37 \\
\midrule
EGMF (GLM3-6B) & \textbf{67.22} & \textbf{65.57} & \textbf{73.97} & \textbf{73.90} & \textbf{87.30} & \textbf{87.09} & \textbf{81.56} & \textbf{81.13} \\
\bottomrule
\end{tabular}
\label{tab:ablation-all}
\end{table*}

\subsection{Ablation Studies}

\textbf{Experimental Design and Overview.}
We conduct comprehensive ablation studies across four benchmark datasets to systematically evaluate the contribution of each component in our EGMF framework. Table~\ref{tab:ablation-all} presents the experimental results, revealing several key insights into the effectiveness of individual modalities and architectural components.

\textbf{Modality Contribution Analysis.}
Our analysis reveals the central importance of textual information in multimodal emotion understanding. Removing the text modality causes the most dramatic performance degradation across all datasets, with drops of 20.44\% on MELD and 16.08\% on MOSEI, confirming text as the primary semantic carrier. While audio and visual modalities show smaller individual contributions (0.5\%-1.5\% improvements), their combined removal leads to more substantial degradation, particularly on Chinese datasets. For instance, removing both audio and visual modalities simultaneously results in a 16.95\% drop on CHERMA compared to only 5.90\% on MELD, suggesting stronger multimodal dependencies in Chinese emotional expressions.

\textbf{Expert Network Component Analysis.}
We examine the individual contribution of each expert network within our multi-scale architecture. The Fine-Grained Local Expert ($E_1$) demonstrates the most significant impact, with its removal causing performance drops ranging from 0.78\% to 1.66\% across datasets. The Global Context Expert ($E_3$) shows comparable importance, particularly on classification tasks like MELD (1.70\% drop) and CHERMA (0.98\% drop). The Semantic Correlation Expert ($E_2$) provides more moderate but consistent contributions across all datasets. This hierarchical importance pattern validates our design rationale that fine-grained local patterns and global contextual information are crucial for effective emotion recognition, while mid-level semantic correlations provide additional refinement.

\textbf{Parameter-Efficient Fine-tuning Analysis.}
Our LoRA-based fine-tuning strategy shows language-specific effectiveness patterns. On English datasets (MELD and MOSEI), LoRA fine-tuning provides consistent improvements of 0.74\%-1.40\%, demonstrating successful adaptation of the pre-trained language model to multimodal emotion tasks. However, we observe performance degradation when applying LoRA to Chinese datasets, likely due to representational mismatches introduced by the English-centric pre-training of the underlying language model. This finding suggests that cross-lingual adaptation strategies require careful consideration of language-specific characteristics.

\textbf{Cross-lingual Modality Synergy and Expert Complementarity.}  
Our cross-lingual analysis shows that Chinese datasets rely more on multimodal fusion, with larger performance drops when audio-visual information is removed, indicating stronger dependence on paralinguistic and visual cues. The expert networks exhibit clear complementarity: $E_1$ captures fine-grained details, $E_2$ models semantic correlations, and $E_3$ encodes global context. Their importance varies across datasets and tasks, with classification benefiting more from $E_1$ and $E_3$, while regression tasks utilize all experts more evenly, demonstrating the framework's adaptability to diverse emotion recognition and sentiment analysis tasks.

\section{Conclusion}

In this paper, we present EGMF, a unified multimodal framework that seamlessly bridges emotion recognition and sentiment analysis through expert-guided feature fusion and large language model integration. The framework employs a multi-scale expert network architecture with three functionally specialized experts and hierarchical dynamic gating mechanisms for adaptive multimodal integration. Through comprehensive evaluation across bilingual datasets (English and Chinese), we demonstrate consistent cross-lingual robustness while revealing universal patterns in multimodal emotional expressions. Our unified design successfully handles both discrete emotion classification and continuous sentiment regression within a single architecture, establishing a new paradigm for multimodal affective computing that provides a foundation for developing more comprehensive emotion understanding systems.

\textbf{Limitations.}  
EGMF performs well on English datasets, but LoRA fine-tuning is less effective on Chinese datasets, highlighting challenges in cross-lingual adaptation. Multimodal reliance varies across datasets, and missing or noisy modalities can affect performance. Although the multi-scale expert network with LLM improves long-range dependency modeling, it may struggle with very long dialogues or rapid emotional shifts. Training costs remain high, and experiments cover only a limited set of emotion labels, which may not fully capture cultural variations in expression.

\bibliography{custom}

@article{1.gkintoni2025neural,
  title={From neural networks to emotional networks: A systematic review of EEG-based emotion recognition in cognitive neuroscience and real-world applications},
  author={Gkintoni, Evgenia and Aroutzidis, Anthimos and Antonopoulou, Hera and Halkiopoulos, Constantinos},
  journal={Brain Sciences},
  volume={15},
  number={3},
  pages={220},
  year={2025}
}

@article{2.yang2025ccin,
  title={CCIN-SA: Composite cross modal interaction network with attention enhancement for multimodal sentiment analysis},
  author={Yang, Li and Zhong, Junhong and Wen, Teng and Liao, Yuan},
  journal={Information Fusion},
  pages={103230},
  year={2025},
  publisher={Elsevier}
}

@article{3.benrouba2023emotional,
  title={Emotional sentiment analysis of social media content for mental health safety},
  author={Benrouba, Ferdaous and Boudour, Rachid},
  journal={Social Network Analysis and Mining},
  volume={13},
  number={1},
  pages={17},
  year={2023},
  publisher={Springer}
}

@article{4.rodriguez2023review,
  title={A review on sentiment analysis from social media platforms},
  author={Rodr{\'\i}guez-Ib{\'a}nez, Margarita and Cas{\'a}nez-Ventura, Antonio and Castej{\'o}n-Mateos, F{\'e}lix and Cuenca-Jim{\'e}nez, Pedro-Manuel},
  journal={Expert Systems with Applications},
  volume={223},
  pages={119862},
  year={2023},
  publisher={Elsevier}
}

@article{5.chen2024evolution,
  title={Evolution and Prospects of Foundation Models: From Large Language Models to Large Multimodal Models.},
  author={Chen, Zheyi and Xu, Liuchang and Zheng, Hongting and Chen, Luyao and Tolba, Amr and Zhao, Liang and Yu, Keping and Feng, Hailin},
  journal={Computers, Materials \& Continua},
  volume={80},
  number={2},
  year={2024}
}

@inproceedings{6.Nguyen_2023,
   title={Conversation Understanding using Relational Temporal Graph Neural Networks with Auxiliary Cross-Modality Interaction},
   url={http://dx.doi.org/10.18653/v1/2023.emnlp-main.937},
   DOI={10.18653/v1/2023.emnlp-main.937},
   booktitle={Proceedings of the 2023 Conference on Empirical Methods in Natural Language Processing},
   publisher={Association for Computational Linguistics},
   author={Nguyen, Cam Van Thi and Mai, Tuan and The, Son and Kieu, Dang and Le, Duc-Trong},
   year={2023},
   pages={15154–15167} }

@article{8.lei2023instructerc,
  title={Instructerc: Reforming emotion recognition in conversation with a retrieval multi-task llms framework},
  author={Lei, Shanglin and Dong, Guanting and Wang, Xiaoping and Wang, Keheng and Wang, Sirui},
  journal={CoRR},
  year={2023}
}

@inproceedings{9.xue2024bioserc,
  title={Bioserc: Integrating biography speakers supported by llms for erc tasks},
  author={Xue, Jieying and Nguyen, Minh-Phuong and Matheny, Blake and Nguyen, Le-Minh},
  booktitle={International Conference on Artificial Neural Networks},
  pages={277--292},
  year={2024},
  organization={Springer}
}

@article{10.zhang2023dialoguellm,
  title={Dialoguellm: Context and emotion knowledge-tuned large language models for emotion recognition in conversations},
  author={Zhang, Yazhou and Wang, Mengyao and Wu, Youxi and Tiwari, Prayag and Li, Qiuchi and Wang, Benyou and Qin, Jing},
  journal={arXiv preprint arXiv:2310.11374},
  year={2023}
}

@INPROCEEDINGS{11.DS-LSTM,
  author={Wang, Jianyou and Xue, Michael and Culhane, Ryan and Diao, Enmao and Ding, Jie and Tarokh, Vahid},
  booktitle={ICASSP 2020 - 2020 IEEE International Conference on Acoustics, Speech and Signal Processing (ICASSP)}, 
  title={Speech Emotion Recognition with Dual-Sequence LSTM Architecture}, 
  year={2020},
  volume={},
  number={},
  pages={6474-6478},
  doi={10.1109/ICASSP40776.2020.9054629}}

@article{12.ghosal2019dialoguegcn,
  title={Dialoguegcn: A graph convolutional neural network for emotion recognition in conversation},
  author={Ghosal, Deepanway and Majumder, Navonil and Poria, Soujanya and Chhaya, Niyati and Gelbukh, Alexander},
  journal={arXiv preprint arXiv:1908.11540},
  year={2019}
}

@inproceedings{13.hazarika2018conversational,
  title={Conversational memory network for emotion recognition in dyadic dialogue videos},
  author={Hazarika, Devamanyu and Poria, Soujanya and Zadeh, Amir and Cambria, Erik and Morency, Louis-Philippe and Zimmermann, Roger},
  booktitle={Proceedings of the conference. Association for Computational Linguistics. North American Chapter. Meeting},
  volume={2018},
  pages={2122},
  year={2018}
}

@article{14.zhang2023dual,
  title={A dual-direction attention mixed feature network for facial expression recognition},
  author={Zhang, Saining and Zhang, Yuhang and Zhang, Ye and Wang, Yufei and Song, Zhigang},
  journal={Electronics},
  volume={12},
  number={17},
  pages={3595},
  year={2023},
  publisher={MDPI}
}

@article{15.kim2021emoberta,
  title={Emoberta: Speaker-aware emotion recognition in conversation with roberta},
  author={Kim, Taewoon and Vossen, Piek},
  journal={arXiv preprint arXiv:2108.12009},
  year={2021}
}

@article{16.liu2019roberta,
  title={Roberta: A robustly optimized bert pretraining approach},
  author={Liu, Yinhan and Ott, Myle and Goyal, Naman and Du, Jingfei and Joshi, Mandar and Chen, Danqi and Levy, Omer and Lewis, Mike and Zettlemoyer, Luke and Stoyanov, Veselin},
  journal={arXiv preprint arXiv:1907.11692},
  year={2019}
}

@article{17.li2022emocaps,
  title={EmoCaps: Emotion capsule based model for conversational emotion recognition},
  author={Li, Zaijing and Tang, Fengxiao and Zhao, Ming and Zhu, Yusen},
  journal={arXiv preprint arXiv:2203.13504},
  year={2022}
}

@inproceedings{18.qin2023bert,
  title={Bert-erc: Fine-tuning bert is enough for emotion recognition in conversation},
  author={Qin, Xiangyu and Wu, Zhiyu and Zhang, Tingting and Li, Yanran and Luan, Jian and Wang, Bin and Wang, Li and Cui, Jinshi},
  booktitle={Proceedings of the AAAI conference on artificial intelligence},
  volume={37},
  number={11},
  pages={13492--13500},
  year={2023}
}

@inproceedings{19.tengstrand2014eacl,
  title={EACL-Expansion of Abbreviations in CLinical text},
  author={Tengstrand, Lisa and Megyesi, Be{\'a}ta and Henriksson, Aron and Duneld, Martin and Kvist, Maria},
  booktitle={Proceedings of the 3rd Workshop on Predicting and Improving Text Readability for Target Reader Populations (PITR)},
  pages={94--103},
  year={2014}
}

@inproceedings{22.hazarika2020misa,
  title={Misa: Modality-invariant and-specific representations for multimodal sentiment analysis},
  author={Hazarika, Devamanyu and Zimmermann, Roger and Poria, Soujanya},
  booktitle={Proceedings of the 28th ACM international conference on multimedia},
  pages={1122--1131},
  year={2020}
}

@inproceedings{23.rahman2020integrating,
  title={Integrating multimodal information in large pretrained transformers},
  author={Rahman, Wasifur and Hasan, Md Kamrul and Lee, Sangwu and Zadeh, Amir and Mao, Chengfeng and Morency, Louis-Philippe and Hoque, Ehsan},
  booktitle={Proceedings of the conference. Association for computational linguistics. Meeting},
  volume={2020},
  pages={2359},
  year={2020}
}

@article{24.hu2021dialoguecrn,
  title={Dialoguecrn: Contextual reasoning networks for emotion recognition in conversations},
  author={Hu, Dou and Wei, Lingwei and Huai, Xiaoyong},
  journal={arXiv preprint arXiv:2106.01978},
  year={2021}
}

@article{26.shen2021directed,
  title={Directed acyclic graph network for conversational emotion recognition},
  author={Shen, Weizhou and Wu, Siyue and Yang, Yunyi and Quan, Xiaojun},
  journal={arXiv preprint arXiv:2105.12907},
  year={2021}
}

@article{27.ai2024gcn,
  title={Der-gcn: Dialog and event relation-aware graph convolutional neural network for multimodal dialog emotion recognition},
  author={Ai, Wei and Shou, Yuntao and Meng, Tao and Li, Keqin},
  journal={IEEE Transactions on Neural Networks and Learning Systems},
  volume={36},
  number={3},
  pages={4908--4921},
  year={2024},
  publisher={IEEE}
}

@inproceedings{28.poria-etal-2019-meld,
    title = "{MELD}: A Multimodal Multi-Party Dataset for Emotion Recognition in Conversations",
    author = "Poria, Soujanya  and
      Hazarika, Devamanyu  and
      Majumder, Navonil  and
      Naik, Gautam  and
      Cambria, Erik  and
      Mihalcea, Rada",
    editor = "Korhonen, Anna  and
      Traum, David  and
      M{\`a}rquez, Llu{\'i}s",
    booktitle = "Proceedings of the 57th Annual Meeting of the Association for Computational Linguistics",
    month = jul,
    year = "2019",
    address = "Florence, Italy",
    publisher = "Association for Computational Linguistics",
    url = "https://aclanthology.org/P19-1050/",
    doi = "10.18653/v1/P19-1050",
    pages = "527--536"
}

@inproceedings{29.sun-etal-2023-layer,
    title = "Layer-wise Fusion with Modality Independence Modeling for Multi-modal Emotion Recognition",
    author = "Sun, Jun  and
      Han, Shoukang  and
      Ruan, Yu-Ping  and
      Zhang, Xiaoning  and
      Zheng, Shu-Kai  and
      Liu, Yulong  and
      Huang, Yuxin  and
      Li, Taihao",
    editor = "Rogers, Anna  and
      Boyd-Graber, Jordan  and
      Okazaki, Naoaki",
    booktitle = "Proceedings of the 61st Annual Meeting of the Association for Computational Linguistics (Volume 1: Long Papers)",
    month = jul,
    year = "2023",
    address = "Toronto, Canada",
    publisher = "Association for Computational Linguistics",
    url = "https://aclanthology.org/2023.acl-long.39/",
    doi = "10.18653/v1/2023.acl-long.39",
    pages = "658--670"
}

@misc{30.liu2022make,
    title={Make Acoustic and Visual Cues Matter: CH-SIMS v2.0 Dataset and AV-Mixup Consistent Module}, 
    author={Yihe Liu and Ziqi Yuan and Huisheng Mao and Zhiyun Liang and Wanqiuyue Yang and Yuanzhe Qiu and Tie Cheng and Xiaoteng Li and Hua Xu and Kai Gao},
    year={2022},
    eprint={2209.02604},
    archivePrefix={arXiv},
    primaryClass={cs.MM}
  }

@inproceedings{31.bagher-zadeh-etal-2018-multimodal,
    title = "Multimodal Language Analysis in the Wild: {CMU}-{MOSEI} Dataset and Interpretable Dynamic Fusion Graph",
    author = "Bagher Zadeh, AmirAli  and
      Liang, Paul Pu  and
      Poria, Soujanya  and
      Cambria, Erik  and
      Morency, Louis-Philippe",
    editor = "Gurevych, Iryna  and
      Miyao, Yusuke",
    booktitle = "Proceedings of the 56th Annual Meeting of the Association for Computational Linguistics (Volume 1: Long Papers)",
    month = jul,
    year = "2018",
    address = "Melbourne, Australia",
    publisher = "Association for Computational Linguistics",
    url = "https://aclanthology.org/P18-1208/",
    doi = "10.18653/v1/P18-1208",
    pages = "2236--2246"
}

@article{50.zadeh2017tensor,
  title={Tensor fusion network for multimodal sentiment analysis},
  author={Zadeh, Amir and Chen, Minghai and Poria, Soujanya and Cambria, Erik and Morency, Louis-Philippe},
  journal={arXiv preprint arXiv:1707.07250},
  year={2017}
}

@article{51.liu2018efficient,
  title={Efficient low-rank multimodal fusion with modality-specific factors},
  author={Liu, Zhun and Shen, Ying and Lakshminarasimhan, Varun Bharadhwaj and Liang, Paul Pu and Zadeh, Amir and Morency, Louis-Philippe},
  journal={arXiv preprint arXiv:1806.00064},
  year={2018}
}

@inproceedings{52.rahman2020integrating,
  title={Integrating multimodal information in large pretrained transformers},
  author={Rahman, Wasifur and Hasan, Md Kamrul and Lee, Sangwu and Zadeh, Amir and Mao, Chengfeng and Morency, Louis-Philippe and Hoque, Ehsan},
  booktitle={Proceedings of the conference. Association for computational linguistics. Meeting},
  volume={2020},
  pages={2359},
  year={2020}
}

@inproceedings{53.tsai2019multimodal,
  title={Multimodal transformer for unaligned multimodal language sequences},
  author={Tsai, Yao-Hung Hubert and Bai, Shaojie and Liang, Paul Pu and Kolter, J Zico and Morency, Louis-Philippe and Salakhutdinov, Ruslan},
  booktitle={Proceedings of the conference. Association for computational linguistics. Meeting},
  volume={2019},
  pages={6558},
  year={2019}
}

@inproceedings{55.yu2021learning,
  title={Learning modality-specific representations with self-supervised multi-task learning for multimodal sentiment analysis},
  author={Yu, Wenmeng and Xu, Hua and Yuan, Ziqi and Wu, Jiele},
  booktitle={Proceedings of the AAAI conference on artificial intelligence},
  volume={35},
  number={12},
  pages={10790--10797},
  year={2021}
}

@inproceedings{56.zhang2022wenetspeech,
  title={Wenetspeech: A 10000+ hours multi-domain mandarin corpus for speech recognition},
  author={Zhang, Binbin and Lv, Hang and Guo, Pengcheng and Shao, Qijie and Yang, Chao and Xie, Lei and Xu, Xin and Bu, Hui and Chen, Xiaoyu and Zeng, Chenchen and others},
  booktitle={ICASSP 2022-2022 IEEE International Conference on Acoustics, Speech and Signal Processing (ICASSP)},
  pages={6182--6186},
  year={2022},
  organization={IEEE}
}

@article{57.hu2022unimse,
  title={UniMSE: Towards unified multimodal sentiment analysis and emotion recognition},
  author={Hu, Guimin and Lin, Ting-En and Zhao, Yi and Lu, Guangming and Wu, Yuchuan and Li, Yongbin},
  journal={arXiv preprint arXiv:2211.11256},
  year={2022}
}

@inproceedings{58.li2023unisa,
  title={Unisa: Unified generative framework for sentiment analysis},
  author={Li, Zaijing and Lin, Ting-En and Wu, Yuchuan and Liu, Meng and Tang, Fengxiao and Zhao, Ming and Li, Yongbin},
  booktitle={Proceedings of the 31st ACM international conference on multimedia},
  pages={6132--6142},
  year={2023}
}

@inproceedings{59.lv2021progressive,
  title={Progressive modality reinforcement for human multimodal emotion recognition from unaligned multimodal sequences},
  author={Lv, Fengmao and Chen, Xiang and Huang, Yanyong and Duan, Lixin and Lin, Guosheng},
  booktitle={Proceedings of the IEEE/CVF conference on computer vision and pattern recognition},
  pages={2554--2562},
  year={2021}
}

@article{60.li2023ga2mif,
  title={GA2MIF: Graph and attention based two-stage multi-source information fusion for conversational emotion detection},
  author={Li, Jiang and Wang, Xiaoping and Lv, Guoqing and Zeng, Zhigang},
  journal={IEEE Transactions on affective computing},
  volume={15},
  number={1},
  pages={130--143},
  year={2023},
  publisher={IEEE}
}

@inproceedings{99.eyben2010opensmile,
  title={Opensmile: the munich versatile and fast open-source audio feature extractor},
  author={Eyben, Florian and W{\"o}llmer, Martin and Schuller, Bj{\"o}rn},
  booktitle={Proceedings of the 18th ACM international conference on Multimedia},
  pages={1459--1462},
  year={2010}
}

@inproceedings{100.tao2021someone,
  title={Is someone speaking? exploring long-term temporal features for audio-visual active speaker detection},
  author={Tao, Ruijie and Pan, Zexu and Das, Rohan Kumar and Qian, Xinyuan and Shou, Mike Zheng and Li, Haizhou},
  booktitle={Proceedings of the 29th ACM international conference on multimedia},
  pages={3927--3935},
  year={2021}
}

@inproceedings{101.tan2019efficientnet,
  title={Efficientnet: Rethinking model scaling for convolutional neural networks},
  author={Tan, Mingxing and Le, Quoc},
  booktitle={International conference on machine learning},
  pages={6105--6114},
  year={2019},
  organization={PMLR}
}

@article{102.zhang2016joint,
  title={Joint face detection and alignment using multitask cascaded convolutional networks},
  author={Zhang, Kaipeng and Zhang, Zhanpeng and Li, Zhifeng and Qiao, Yu},
  journal={IEEE signal processing letters},
  volume={23},
  number={10},
  pages={1499--1503},
  year={2016},
  publisher={IEEE}
}

\appendix

\section{Supplementary Related Work}
\label{appendix:baselines}
This appendix provides detailed discussions of traditional approaches for emotion recognition and sentiment analysis that serve as important baselines in multimodal affective computing research.

\paragraph{RNN-based Methods}For emotion recognition, DS-LSTM \citep{11.DS-LSTM} effectively handles context features at different times and frequencies, capable of separately processing MFCC and Mel spectrograms through specialized modeling of multi-temporal frequency information in speech signals combined with contextual LSTM encoding. DialogueCRN \citep{12.ghosal2019dialoguegcn} introduces cognitive factors into emotion recognition tasks, providing comprehensive understanding of contexts at both contextual and speaker levels through multi-step reasoning mechanisms. 

In sentiment analysis, models such as CMN \citep{13.hazarika2018conversational} and MFN \citep{14.zhang2023dual} utilize LSTM encoders to model word-level and turn-level dependencies. However, RNN-based methods suffer from several limitations, including shallow interaction structures, difficulties in modeling long-term dependencies, and poor parallelization capabilities, making it challenging to adequately capture modal interaction information in long dialogue scenarios.

\paragraph{Transformer-based Methods}EmoBERTa \citep{15.kim2021emoberta} extends RoBERTa \citep{16.liu2019roberta} by adding speaker names before dialogues and inserting separators to predict the current speaker's emotion, thereby enhancing dialogue context modeling capabilities.  BERT-ERC \citep{18.qin2023bert} enhances ERC performance through suggested text, fine-grained classification modules, and two-stage training, demonstrating excellent generalization capabilities. EACL \citep{19.tengstrand2014eacl} is the first ERC model that addresses the emotion similarity problem by incorporating label semantic information, effectively guiding representation learning. 

For sentiment analysis-oriented methods, approaches such as MISA \citep{22.hazarika2020misa} and MAG-BERT \citep{23.rahman2020integrating} introduce early and late fusion mechanisms, jointly modeling multimodal features through Transformer encoding. Despite superior performance over RNN methods, most approaches still rely on manually designed fusion structures or static weighting mechanisms, failing to dynamically adapt to the varying importance of different modalities across different contexts, particularly lacking modeling capabilities for modal temporal associations and speaker state changes in multi-turn dialogues.

\paragraph{Graph Neural Network Methods}DialogueGCN \citep{24.hu2021dialoguecrn} models dialogue utterances as graph nodes with context-based edges, enhancing emotion propagation modeling capabilities by establishing intra-speaker and inter-speaker edges, thereby improving emotion recognition through inter-speaker and self-dependencies.  DAG-ERC \cite{26.shen2021directed} incorporates speaker identity and positional information in Directed Acyclic Graph Neural Networks. DER-GCN \cite{27.ai2024gcn} examines the impact of event relationships on emotions by modeling contextual semantic information and dialogue relationships. While graph structures are beneficial for modeling dialogue relationships, such methods often suffer from fixed graph topological structures, limited modal edge information design, and difficulties in extending to high-dimensional modal fusion scenarios.

\section{Experimental Details}
\label{appendix:B}

\subsection{Feature Extraction}

To maintain experimental consistency and enable fair comparison, we adopt widely-established feature extraction protocols from prior work. Due to variations in dataset characteristics and release timelines, processing methodologies are adapted accordingly for each dataset (see Table \ref{tab:feature_dimensions} for feature dimensionalities):

\textbf{MOSEI}: Visual features are extracted using OpenFace , while audio features are obtained through COVAREP.

\textbf{SIMS-V2}: Following the official dataset configuration, we employ the unaligned version. Visual features are extracted by first localizing faces with TalkNet \cite{100.tao2021someone}, then using OpenFace for feature extraction; audio features are obtained through OpenSMILE \cite{99.eyben2010opensmile}.

\textbf{MELD}: Following the settings of UNIMSE \cite{57.hu2022unimse} and UNISA \cite{58.li2023unisa}. Visual features are extracted using EfficientNet \cite{101.tan2019efficientnet} pre-trained on VGGface2 and AFEW datasets; audio features are obtained by extracting Mel spectrograms through librosa.

\textbf{CHERMA}: Following the dataset publisher's settings. Visual features are extracted by first processing faces with MTCNN \cite{102.zhang2016joint}, then using ResNet-18 for extraction; audio features are obtained using pre-trained wav2vec.

\begin{table}[h]
\centering
\small
\begin{tabular}{lccc}
\toprule
\textbf{Dataset} & \textbf{Text} & \textbf{Audio} & \textbf{Video} \\
\midrule
MELD & 4096 & 64 & 64 \\
CHERMA & 4096 & 1024 & 2048 \\
MOSEI & 4096 & 74 & 35 \\
SIMS-V2 & 4096 & 25 & 177 \\
\bottomrule
\end{tabular}
\caption{Feature Dimensions Across Different Modalities and Datasets.}
\label{tab:feature_dimensions}
\end{table}

\subsection{Experimental Setup}
\begin{table}[h]
\small
\centering
\begin{tabular}{lcccc}
\toprule
\textbf{EGMF} & \textbf{MELD} & \textbf{CHERMA} & \textbf{MOSEI} & \textbf{SIMS-V2} \\
\midrule
Learning Rate & 5e-5 & 5e-6 & 5e-6 & 5e-5 \\
LoRA-r & 8 & - & 8 & - \\
LoRA-alpha & 16 & - & 16 & - \\
Max new tokens & 2 & 2 & 4 & 4 \\
Pseudo tokens & 4 & 4 & 4 & 4 \\
\bottomrule
\end{tabular}
\caption{Hyperparameter settings for EGMF(GLM3-6B).}
\label{tab:hyperparameters}
\end{table}

Table \ref{tab:hyperparameters} shows EGMF (GLM3-6B) hyperparameters across datasets. LoRA fine-tuning is used only for English datasets but not for Chinese datasets , as it demonstrated inferior performance on Chinese multimodal data.
\begin{table}[h]
\small
\centering
\begin{tabular}{lccc}
\toprule
\textbf{Model} & \textbf{LoRA} & \textbf{Trainable} & \textbf{Total Parameters} \\
\midrule
GLM3-6B & 1.95M & 14.02M & 6.26B \\
LLaMA2-7B & 1.05M & 13.17M & 6.74B \\
LLaMA3-8B & 0.85M & 12.98M & 8.03B \\
GLM4-9B & 2.79M & 14.91M & 9.41B \\
\bottomrule
\end{tabular}
\caption{Parameter statistics for LoRA fine-tuning across different base models.}
\label{tab:model_parameters}
\end{table}

Table \ref{tab:model_parameters} presents the parameter efficiency of our LoRA-based fine-tuning approach across different base language models. The LoRA adapters introduce only 0.85M to 2.79M additional parameters, representing less than 0.03\% of the total model parameters. The trainable parameter count remains consistently around 12-15M across all models, demonstrating the scalability of our approach. This parameter efficiency is particularly significant for practical deployment, as it enables effective multimodal emotion recognition adaptation while maintaining computational feasibility and reducing storage requirements for model checkpoints.

\begin{table}[h]
\small
\centering
\begin{tabular}{l p{6cm}}
\toprule
\textbf{Model} & \textbf{Prompt Template} \\
\midrule
\textbf{MELD} & As \textbf{an expert in multimodal emotion recognition}, please analyze the given multimodal content and determine the most appropriate emotional category from the following target set: \textless \textbf{neutral: 0, surprise: 1, fear: 2, sadness: 3, joy: 4, disgust: 5, anger: 6}\textgreater. \\[0.5em]
\addlinespace[0.3em]
\textbf{CHERMA} & As \textbf{an expert in multimodal emotion recognition}, please analyze the provided multimodal content and select the most appropriate category from the following emotion labels: \textless \textbf{anger: 0, disgust: 1, fear: 2, happiness: 3, calm: 4, sadness: 5, surprise: 6} \textgreater. \\[0.5em]
\addlinespace[0.3em]
\textbf{MOSEI} & As \textbf{an expert in multimodal sentiment analysis}, please assess the sentiment intensity of the given multimodal content. The sentiment intensity should be predicted as a real value within the continuous range of \textbf{[-3.0, 3.0]}, where \textbf{-3.0} indicates extremely \textbf{negative sentiment} and \textbf{3.0} indicates extremely \textbf{positive sentiment}. \\[0.5em]
\addlinespace[0.3em]
\textbf{SIMS-V2} & As \textbf{an expert in multimodal sentiment analysis}, please analyze the given multimodal input and predict the sentiment intensity as a real-valued score within the continuous range \textbf{[-1.0, 1.0]}, where \textbf{-1.0} indicates extremely \textbf{negative sentiment} and \textbf{1.0} indicates extremely \textbf{positive sentiment}. \\
\bottomrule
\end{tabular}
\caption{Prompt templates for different multimodal datasets.}
\label{tab:prompt_templates}
\end{table}
As shown in Table~\ref{tab:prompt_templates}, we design prompt templates for four representative multimodal datasets, which reflect two primary paradigms in affective computing. Specifically, MELD and CHERMA datasets employ discrete classification approaches for \textbf{multimodal emotion recognition} tasks using seven predefined emotion categories. In contrast, MOSEI and SIMS-V2 datasets focus on \textbf{multimodal sentiment analysis} through continuous regression, quantifying sentiment intensity within ranges of $[-3.0, 3.0]$ and $[-1.0, 1.0]$, respectively. Notably, all prompt templates adopt expert role positioning strategies, highlighting the importance of role-based prompt engineering in multimodal affective computing.

\section{Per-Category and Cross-Lingual Performance Analysis}
\label{app:detailed_analysis}

\subsection{Per-Class Performance Analysis}

This appendix provides a comprehensive analysis of our model's performance across individual emotion categories, revealing dataset-specific characteristics and cross-lingual patterns in multimodal emotion recognition.

\begin{table}[h]
\footnotesize
\centering
\begin{tabular}{lccccccc}
\toprule
\textbf{neu} & \textbf{sur} & \textbf{fea} & \textbf{sad} & \textbf{joy} & 
\textbf{dis} & \textbf{ang}\\
\midrule
80.37 & 57.86 & 34.67 & 39.62 & 64.08 & 30.30 & 52.98\\
69.06 & 70.63 & 81.19 & 83.22 & 82.71 & 54.65 & 78.31\\
\bottomrule
\end{tabular}
\caption{Per-class F1 scores for MELD (first row) and CHERMA (second row) datasets.}
\label{tab:per_class_f1}
\end{table}

Table \ref{tab:per_class_f1} presents detailed per-class F1 scores, revealing distinct performance patterns across datasets and emotion categories. The MELD dataset demonstrates strong performance on neutral emotions (80.37\%) and moderate success with joy (64.08\%), but exhibits significant challenges in recognizing fear (34.67\%) and disgust (30.30\%). Conversely, the CHERMA dataset shows superior performance on sadness (83.22\%) and joy (82.71\%), with generally more balanced recognition across emotion categories.

\subsection{Cross-Dataset Insights}

The comparative analysis reveals several important findings:
\begin{itemize}[leftmargin=*]
    \item Both datasets consistently struggle with disgust recognition (30.30\% for MELD, 54.65\% for CHERMA), suggesting inherent cross-lingual challenges for this particular emotion category;
    \item Class imbalance significantly impacts model performance, as evidenced by MELD's skewed predictions;
    \item Language and cultural factors may influence emotion expression patterns, contributing to the observed performance differences between English and Chinese datasets.
\end{itemize}

These findings underscore the critical importance of dataset composition and class balance in developing robust multimodal emotion recognition systems, particularly for cross-lingual applications.

\begin{figure}[htbp]
\centering
\includegraphics[width=0.78\linewidth]{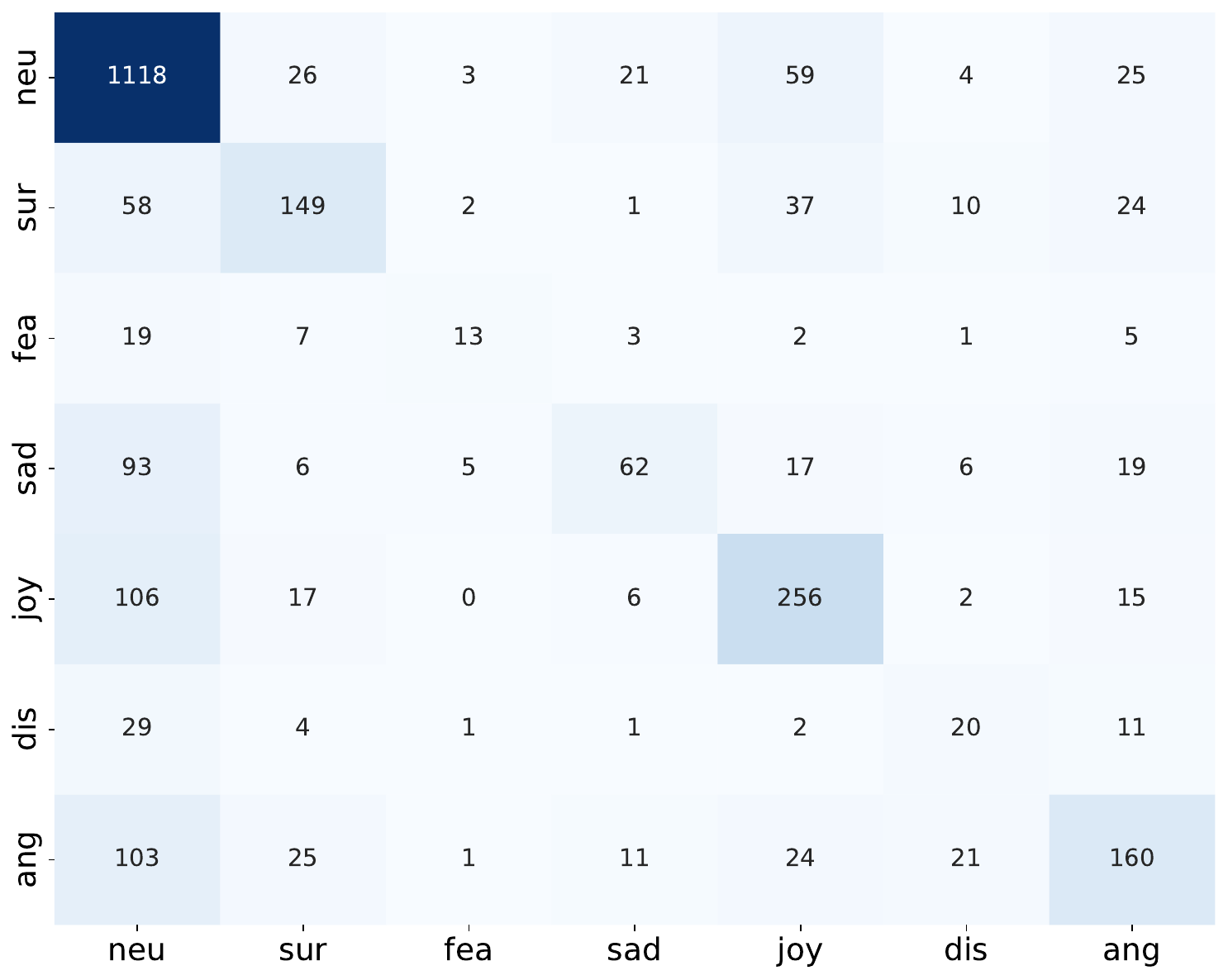}
\caption{Confusion matrix on MELD dataset demonstrating class imbalance challenges in English conversational emotion recognition, with neutral emotions dominating predictions}
\label{fig:meld_conf}
\end{figure}

\begin{figure}[htbp]
\centering
\includegraphics[width=0.78\linewidth]{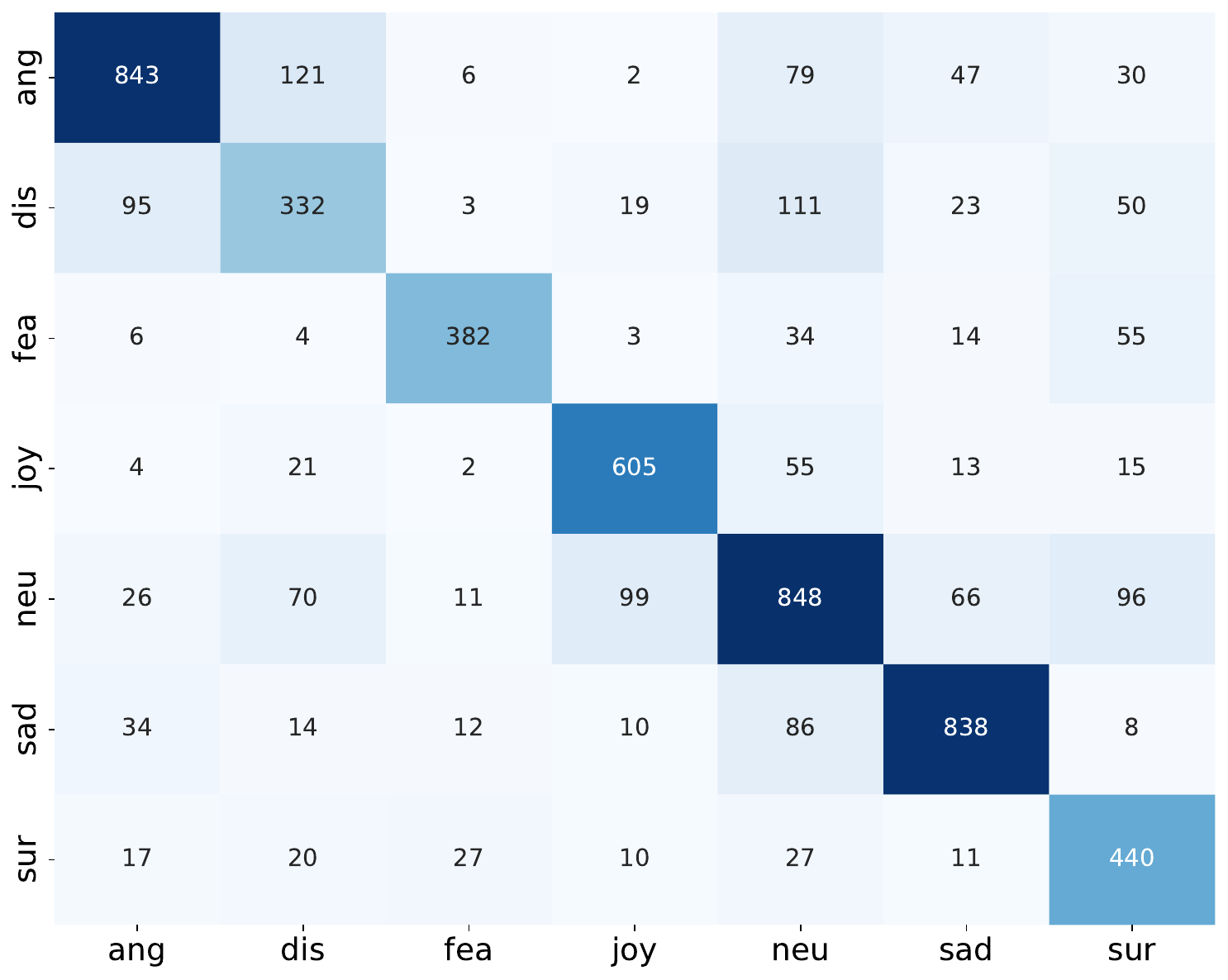}
\caption{Confusion matrix on CHERMA dataset showing more balanced cross-category performance in Chinese multimodal emotion recognition across seven emotion classes}
\label{fig:cherma_conf}
\end{figure}

\begin{figure}[htbp]
\centering
\includegraphics[width=0.78\linewidth]{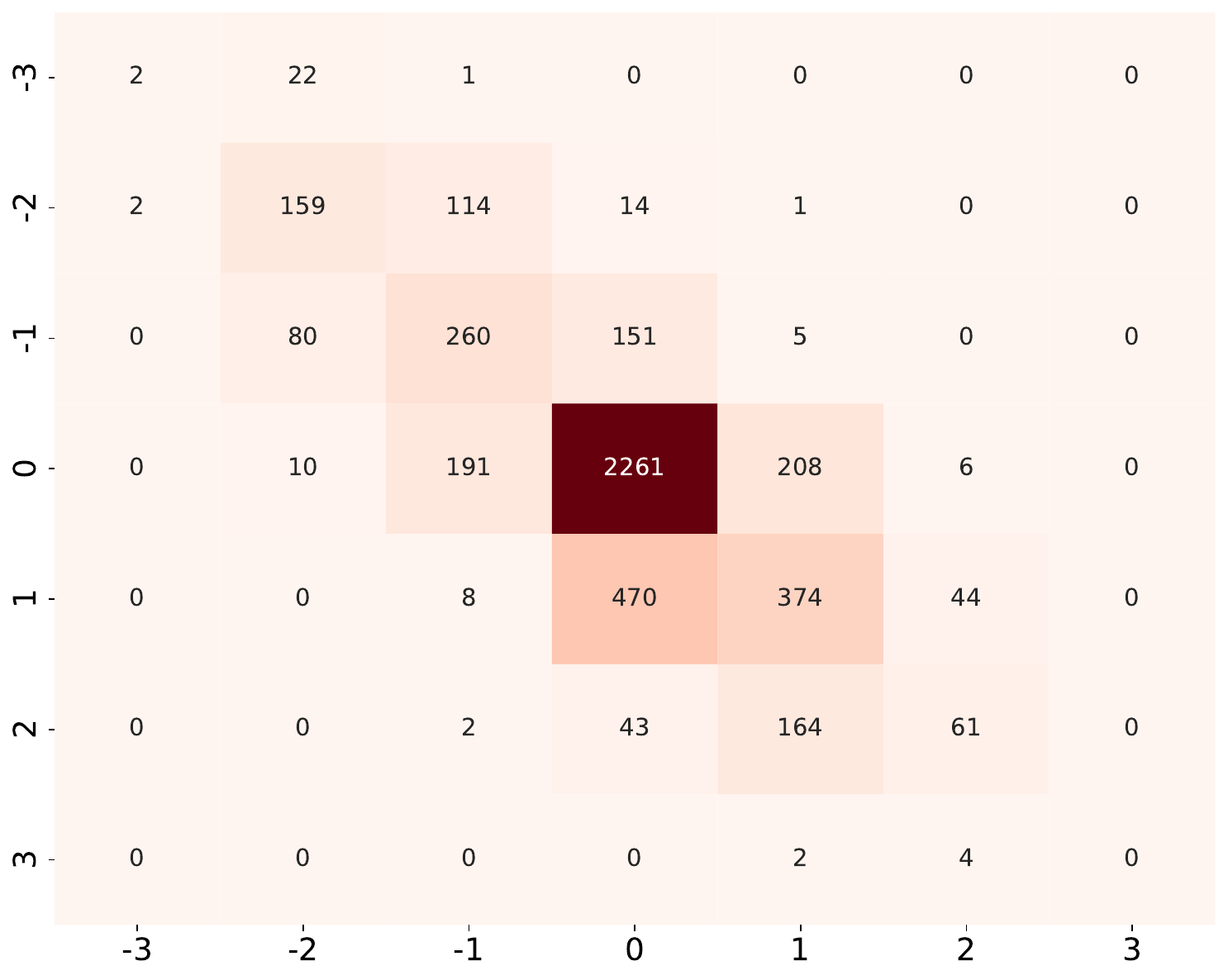}
\caption{Confusion matrix on MOSEI dataset revealing centralization bias in English sentiment intensity regression, with predictions concentrated in moderate ranges despite [-3, 3] scale}
\label{fig:fourth_conf}
\end{figure}

\begin{figure}[htbp]
\centering
\includegraphics[width=0.78\linewidth]{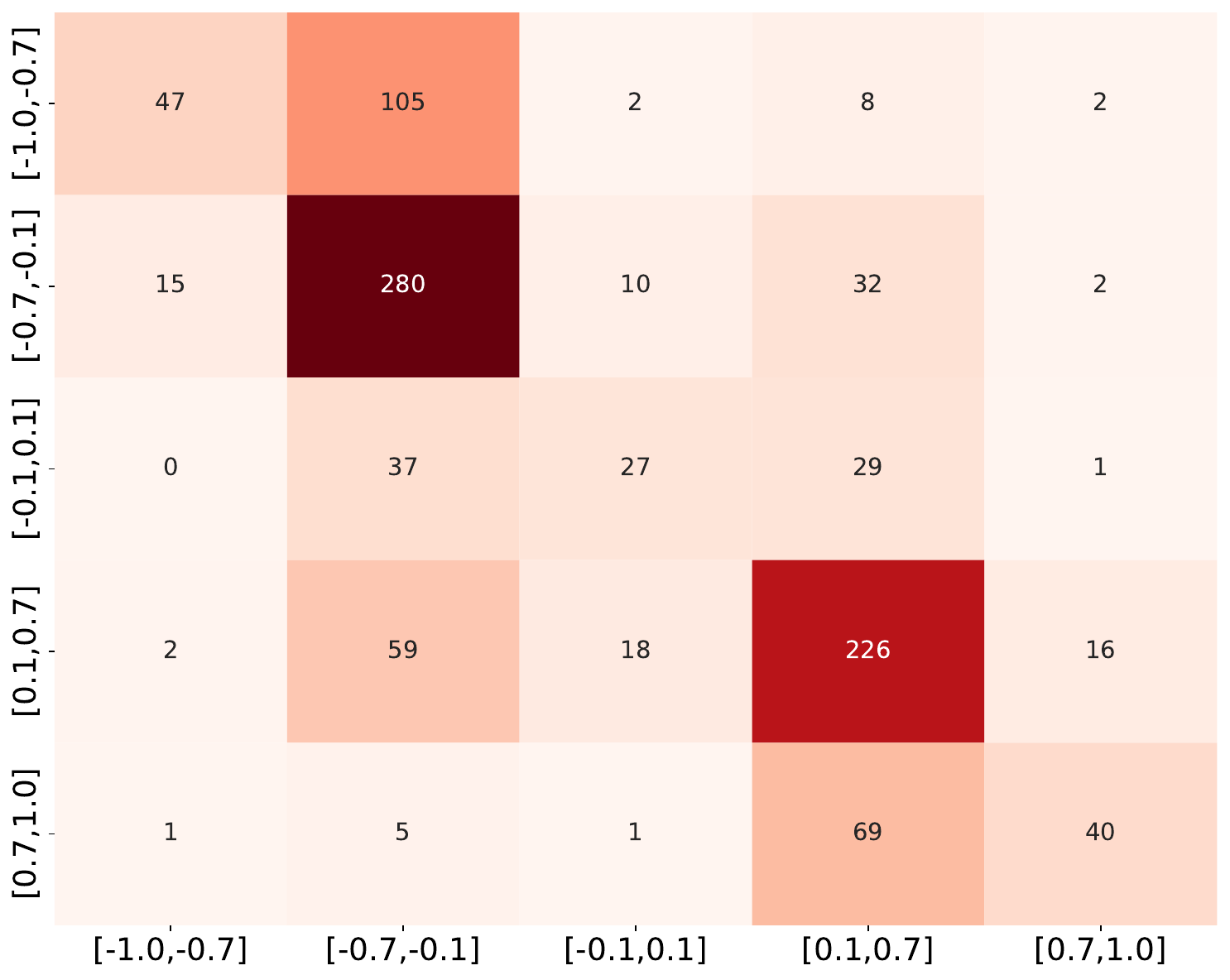}
\caption{Confusion matrix on SIMS-V2 dataset exhibiting improved distribution balance in Chinese sentiment analysis with effective coverage across five discretized intensity intervals}
\label{fig:simsv2_conf}
\end{figure}

\subsection{Confusion Matrix Analysis}

The confusion matrices across the four datasets (Figures \ref{fig:meld_conf}-\ref{fig:simsv2_conf}) reveal distinct patterns in multimodal emotion recognition and sentiment analysis performance, highlighting both task-specific challenges and cross-linguistic differences in affective computing.

\paragraph{Classification Task Performance Analysis} For the MELD dataset, the confusion matrix (Figure \ref{fig:meld_conf}) demonstrates a significant class imbalance issue, with neutral emotions achieving the highest recognition accuracy (diagonal value: 1118), while fear and disgust show substantially lower performance with frequent misclassifications into neutral categories. This pattern aligns with our quantitative F1 scores, where neutral emotions achieved 80.37\% F1-score while fear and disgust obtained only 34.67\% and 30.30\%, respectively. The systematic bias toward neutral predictions suggests that the model struggles to distinguish subtle emotional expressions from baseline states in English conversational contexts.

In contrast, the CHERMA dataset (Figure \ref{fig:cherma_conf}) exhibits more balanced performance across emotion categories, with anger (843), sadness (838), and neutral (848) showing relatively comparable diagonal values. The improved balance indicates that Chinese emotional expressions may possess more distinctive multimodal features, facilitating better cross-category discrimination. Notably, disgust remains challenging across both languages (diagonal values: 33 for CHERMA vs. minimal for MELD), suggesting universal recognition difficulties for this particular emotion category.

\paragraph{Regression Task Characteristics} The MOSEI dataset's confusion matrix (Figure \ref{fig:fourth_conf}) reveals a pronounced centralization bias, with the majority of predictions concentrated in the [-1, 1] sentiment range despite the theoretical [-3, 3] scale. This pattern indicates that the model exhibits conservative behavior in extreme sentiment prediction, potentially due to insufficient training samples in high-intensity sentiment regions or inherent difficulty in fine-grained sentiment intensity estimation. The sparse predictions in extreme categories ($\pm$2, $\pm$3) highlight the challenge of distinguishing subtle gradations in sentiment intensity.

The SIMS-V2 dataset (Figure \ref{fig:simsv2_conf}) demonstrates more balanced distribution across the five discretized sentiment intervals, with stronger performance in negative sentiment regions ([-1.0, -0.7] and [-0.7, -0.1]). The improved balance compared to MOSEI suggests that the narrower sentiment range [-1, 1] may be more suitable for practical sentiment analysis applications, reducing the complexity of fine-grained intensity discrimination.

\paragraph{Cross-Linguistic Error Pattern Analysis} A systematic comparison of the confusion matrices (Figures \ref{fig:meld_conf} and \ref{fig:fourth_conf} vs. Figures \ref{fig:cherma_conf} and \ref{fig:simsv2_conf}) reveals language-specific error patterns. English datasets (MELD, MOSEI) exhibit stronger central tendency bias, with frequent misclassifications toward neutral/mild categories. This phenomenon may reflect cultural differences in emotional expression intensity or dataset construction methodologies. Chinese datasets (CHERMA, SIMS-V2) show more distributed error patterns, suggesting that Chinese multimodal emotional expressions may provide clearer discriminative signals across different affective states.

These findings collectively demonstrate that while our EGMF framework achieves competitive performance across diverse multimodal affective computing tasks, specific improvements are needed for extreme emotion recognition, cross-linguistic adaptation, and fine-grained sentiment intensity estimation to enhance practical applicability in real-world scenarios.

\end{document}